# Paraconsistent-Lib: an intuitive PAL2v algorithm Python Library

**Arnaldo de Carvalho Junior**
Embedded Artificial Intelligence Laboratory (EAILab)
Federal Institute of Sao Paulo (IFSP)
Sao Paulo, BR
adecarvalhojr@ifsp.edu.br

**Diego Oliveira da Cruz**
Embedded Artificial Intelligence Laboratory (EAILab)
Federal Institute of Sao Paulo (IFSP)
Sao Paulo, BR
cruz.diego@aluno.ifsp.edu.br

**Bruno da Silva Alves**
Embedded Artificial Intelligence Laboratory (EAILab)
Federal Institute of Sao Paulo (IFSP)
Sao Paulo, BR
alves.bruno1@aluno.ifsp.edu.br

**Fernando da Silva Paulo Junior**
Embedded Artificial Intelligence Laboratory (EAILab)
Federal Institute of Sao Paulo (IFSP)
Sao Paulo, BR
fernando.junior1@aluno.ifsp.edu.br

**João Inacio da Silva Filho**
Applied Paraconsistent Logic Group
Santa Cecilia University (UNISANTA)
Sao Paulo, BR
inacio@unisanta.br

**Abstract:**

This paper introduces Paraconsistent-Lib, an open-source, easy-to-use Python library for building PAL2v algorithms in reasoning and decision-making systems. Paraconsistent-Lib is designed as a general-purpose library of PAL2v standard calculations, presenting three types of results: paraconsistent analysis in one of the 12 classical lattice PAL2v regions, paraconsistent analysis node (PAN) outputs, and a decision output. With Paraconsistent-Lib, well-known PAL2v algorithms such as Para-analyzer, ParaExtrCTX, PAL2v Filter, paraconsistent analysis network (PANnet), and paraconsistent neural network (PNN) can be written in stand-alone or network form, reducing complexity, code size, and bugs, as two examples presented in this paper. Given its stable state, Paraconsistent-Lib is an active development to respond to user-required features and enhancements received on GitHub.

## 1. Introduction

The desire to create an automaton capable of imitating human behavior is long-standing. In ancient Greece, Philon of Byzantium developed an automaton in the form of a life-size maid to serve wine and water mixed in exact proportions [Vasileiadou and Kalligeropoulos, 2007].

Since decision-making is an integral part of human life, an effort to develop machines capable of not only imitating, but also reasoning and solving problems has generated interest in the philosophy, science, engineering and media. Technological evolution allowed the search for systems capable of imitating human reasoning to gain attention after the mid-20th century [Foote, 2022].

Artificial intelligence (AI) brings together all approaches and tools that add the capacity of computer systems to perform tasks associated with human intelligence, such as learning, reasoning, problem solving, perception, identification, analysis, association, diagnosis and decision making [Jiang et al., 2022].

Classical formal logic is one of the pillars of computational systems and, by extension, one of the fundamental tools used in the development of AI algorithms [Thomason, 2024]. This logic does not accept contradiction. But the real world does not work like that.

Contradictions appear as different fields of science evolve and become more complex. Since Heraclitus in ancient Greece, several philosophers have dedicated efforts to the study of contradiction, which allowed the emergence of several non classical logic, such as Fuzzy, Dynamic Semantics, Quantum, and Paraconsistent Logic (PL) [De Carvalho Junior et al., 2024].

PL includes inconsistencies in its structure, allowing contradictory situations to be handled without incurring triviality [Da Silva Filho et al., 2011]. Paraconsistent Annotated Logic by 2-value annotation (PAL2v), sometimes also referred to as Paraconsistent Annotated Evidential logic (PAL $\varepsilon\tau$), makes use of two values, or degrees of evidence, to better express knowledge about a proposition [Da Silva Filho, 2012]. Researchers' interest in PAL2v to develop systems capable of dealing with conflicting, contradictory, uncertain, or noise-contaminated data has increased. PAL2v has been used in several areas of human reasoning, such as biology, logistics, statistical process control, robotics, expert systems, and AI [De Carvalho Junior et al., 2024].

Algorithms built with PAL2v have been used in a wide variety of applications, such as for the characterization of skin cancer lesions [Garcia et al., 2019] and choosing the best route in data communication systems [Da Silva Filho et al., 2021]. Artificial neural networks (ANN) built with PAL2v neurons, or paraconsistent neural networks (PNN), were used for identification [De Carvalho Junior et al., 2021] and control of rotating inverted pendulum systems [De Carvalho Junior et al., 2023], and for the Detection of nitrogen oxide emissions (NOx) in Petrochemical Combustion Systems [Rodrigues et al., 2024]. Recently, a new technology for gesture recognition using surface electromyography (sEMG) signals designed with random forest algorithms using PAL2v (Paraconsistent Random Forest - PRF) [Favieiro and Balbinot, 2019] showed promising results, compared to other techniques [Favieiro et al., 2025].

Here we propose and describe Paraconsistent-Lib, a userfriendly Python library to support the development of PAL2v reasoning, the outputs, and examples of application. The paper is structured as follows: In Section 2, we provide a review of PAL2v, while in Section 3, we describe the implementation details of the Paraconsistent-Lib. In Section 4, we describe the two examples of the application of Paraconsistent-Lib. Finally, in Section 5, we draw some final remarks and provide insights into future developments of the library.

## 2. PAL2v Review

PL receives the term "annotated" (PAL) when the evidence of a proposition $P$ is represented by annotation through a lattice of four vertices [Da Silva Filho, 2012]. The four extreme logic states indicated at the vertices of the PAL lattice are True (t), False (F), Paracomplete ($\bot$), and Inconsistent ($\top$) [De Carvalho Junior et al, 2024].

PAL2v uses an ordained pair of values ($\mu_1$, $\mu_2$), normalized between [0,1], to deliver greater precision in representing the evidence of a proposition $P$ in a lattice diagram. The first input, $\mu$, (or $\mu_1$) represents the degree of favorable evidence. The complement of the second input ($\mu_2$), is named the degree of unfavorable evidence ($\lambda$), calculated by Eq. (1).

$$\lambda = 1 - \mu_2 \qquad (1)$$

Through transformations in a Unitary Quadrant of the Cartesian Plane (UQCP) and some algebraic interpretations, as well explained in [Da Silva Filho, 2012], The PAL2v result in coordinate form $\epsilon_\tau = (D_C, D_{CT})$ can be obtained, according to Eq. (2) and Eq (3). Fig. 1 presents the projection of $\mu$ and $\lambda$ values inside the PAL2v lattice, the horizontal ($D_C$) and vertical ($D_{CT}$) axes, and $\varepsilon\tau(D_C, D_{CT})$ output.

$$D_C = \mu - \lambda \qquad (2)$$

$$D_{CT} = \mu + \lambda - 1 \qquad (3)$$

Where $D_C$ is the certainty degree, and $D_{CT}$ is the contradiction degree, with values in the interval of [-1,1]. The range of certainty values in which the $D_C$ can vary without being limited by $D_{CT}$, called certainty interval ($\varphi$) is calculated by Eq. (4).

$$\varphi = 1 - |D_{CT}| \tag{4}$$

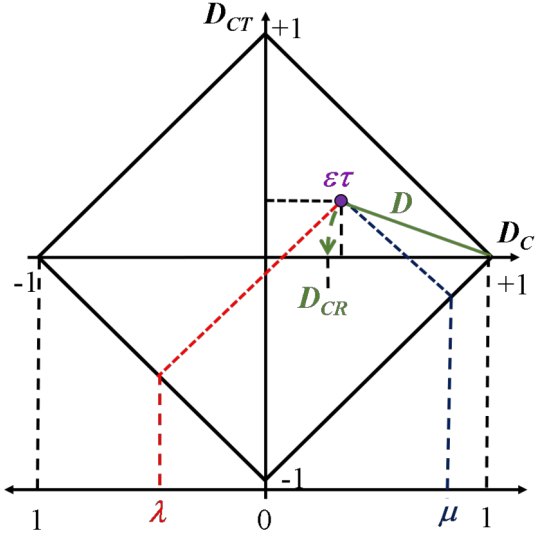

**Figure 1:** PAL2v Lattice.

A single PAL2v output can be obtained by reducing the contradictions to calculate the projection of vector ($D$) on the $Dc$ axis, known as the real certainty degree ($D_{CR}$). This can be calculated according to Eqs. (5) to (7). The center of the PAL2v lattice, where both $Dc$ and $D_{CT}$ are equal to zero is called "undefined".

$$d = \sqrt{(1-|D_C|)^2 + D_{CT}^2} \tag{5}$$

$$D = \begin{cases} 1 \rightarrow d > 1 \\ d \rightarrow d \leq 1 \end{cases} \tag{6}$$

$$D_{CR} = \begin{cases} 1 - D \rightarrow D_C > 0 \\ D - 1 \rightarrow D_C < 0 \\ 0 \rightarrow D_C = 0 \end{cases} \tag{7}$$

In the Para-analyzer algorithm, the PAL2v diagram divided by regions, called paraconsistent logical states. The most common application is to consider 12 logical states. This algorithm can be used for classification or in the decision-making process, as used to decide the direction of robots [De Carvalho Junior et al, 2024]. Fig. 2 presents the four maximum corners (t, F, ⊤, ⊥) and the internal eight logical states.

More complex algorithms can be designed, such as the $ParaExtr_{CTX}$, PAL2v Filter, paraconsistent analysis network (PANnet) and paraconsistent neural networks (PNN) if the PAL2v results are normalized in the interval [0,1], as Eqs. (8) to (11), and built in form of a structured code, called paraconsistent analysis node (PAN) [De Carvalho Junior et al, 2024]. Fig. 3 presents the PAN symbol.

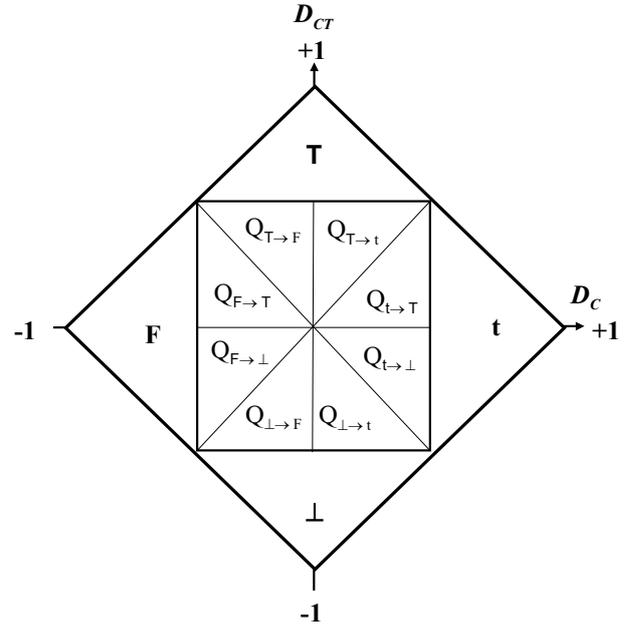

F → False
t → True
⊤ → Inconsistent
⊥ → Indeterminate or Paracomplete
$Q_{(t \rightarrow \top)}$ → Quasi-true, tending to inconsistent
$Q_{(t \rightarrow \bot)}$ → Quasi-true, tending to paracomplete
$Q_{(F \rightarrow \top)}$ → Quasi-false, tending to inconsistent
$Q_{(F \rightarrow \bot)}$ → Quasi-false, tending to paracomplete
$Q_{(\top \rightarrow t)}$ → Quasi-inconsistent, tending to true
$Q_{(\top \rightarrow F)}$ → Quasi-inconsistent, tending to false
$Q_{(\bot \rightarrow t)}$ → Quasi-paracomplete, tending to true
$Q_{(\bot \rightarrow F)}$ → Quasi-paracomplete, tending to false

**Figure 2:** PAL2v lattice with 12 logical states used by the Para-analyzer algorithm.

$$\mu_E = \frac{D_C + 1}{2} \tag{8}$$

$$\mu_{ECT} = \frac{D_{CT}+1}{2} \qquad (9)$$

$$\mu_{ER} = \frac{D_{CR}+1}{2} \qquad (10)$$

$$\varphi_E = 1 - |2\mu_{ECT} - 1| \qquad (11)$$

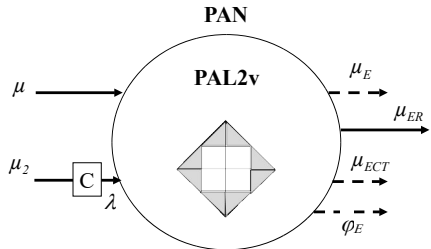

**Figure 3:** PAN symbol.

Paraconsistent artificial neural cells (PANC) can be built by adding rules and equations to the basic PAN. The most common are the standard paraconsistent artificial neural cell ($PANC_S$), paraconsistent neural cell of learning ($PANC_L$), the cell of learning by contradiction extraction ($PANCL_{CTX}$) and the decision cell ($PANC_D$) [Da Silva Filho 2012, De Carvalho Junior et al, 2024].

One direct application of the PAN is the $ParaExtr_{CTX}$ [Da Silva Filho et al, 2011]. It consists of a recurrent analysis of a numerical vector dataset by a PAN, as the flowchart presented in Figure 4. This algorithm works according to the following steps:

1. First, the dataset is normalized between [0,1].
2. Take the highest ($\mu max$) and lowest ($\mu min$) values of the dataset.
3. Let $\mu = \mu max$. Let $\lambda = \mu min$.
4. Calculate $\mu_{ER}$.
5. Remove from the dataset the values used in the PAN analysis.
6. Add the current $\mu_{ER}$ to the dataset.
7. Return to step 2 until only one value remains in the dataset, which will be the final output of the $ParaExtr_{CTX}$ analysis.

$ParaExtr_{CTX}$ is good to extract contradictions between samples of the same measurement or variable [De Carvalho Junior et al, 2024].

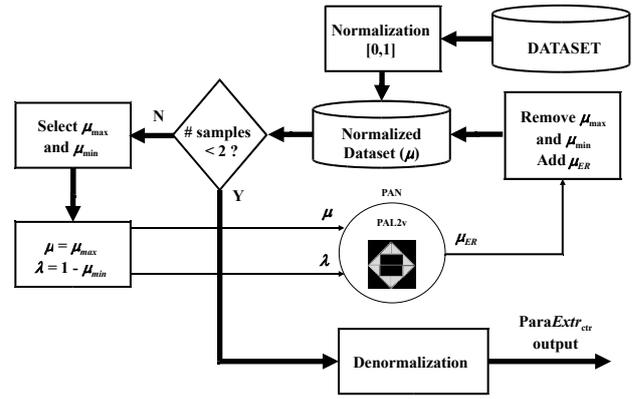

**Figure 4:** $ParaExtr_{CTX}$ flowchart.

## 3. Paraconsistent-Lib Overview

Paraconsistent-Lib implements PAL2v rules and equations, enabling the modeling of reasoning systems that naturally deal with inconsistencies and uncertainties. The library offers a modular architecture based on pluggable logic blocks, facilitating the composition of complex inference networks [Da Cruz; Alves; Carvalho Junior, 2025]. It was designed and tested in Python 3.9 or superior releases, and can be installed as:

```
pip install paraconsistent
```

Paraconsistent-Lib requires just three inputs: $\mu$ and $\lambda$ values and a factor of control called $Ft_C$. The default value is 0.5. This variable is involved in 2 key aspects.

1. Changes the geometry areas of the 12 logical states. FtC higher than 0.5 reduces the areas of the logical states t and F and increases the areas of ⊤ and ⊥, as shown in Fig. 5a. FtC lower than 0.5 reduces the areas of the logical states ⊤ and ⊥ and increases the areas of t and F, as shown in Fig. 5b.
2. FtC serves as a threshold value for the PANCD output (S1), as Eq. (12).

$$S_1 = \begin{cases} True \rightarrow \mu_{ER} > FtC \\ False \rightarrow \mu_{ER} < FtC \\ 0.5 \rightarrow \mu_{ER} = FtC \end{cases} \qquad (12)$$

In the current version, Paraconsistent-Lib can deliver outputs as a Para-analyzer, offering the results of the 12 states, a PAN offering all the outputs calculated as the Eqs. (1) to (11) or a PANCd. It is important to use the correct notation for inputs and outputs when using the Paraconsistent-Lib, as presented in Table 1.

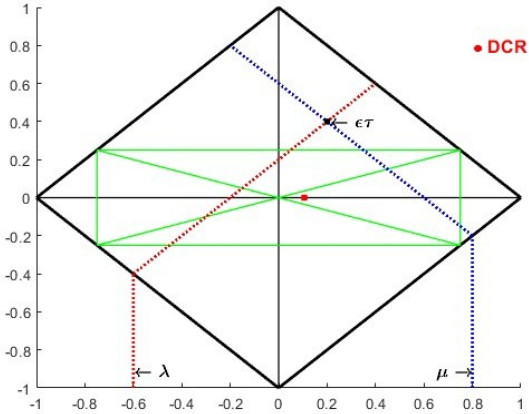

(a)

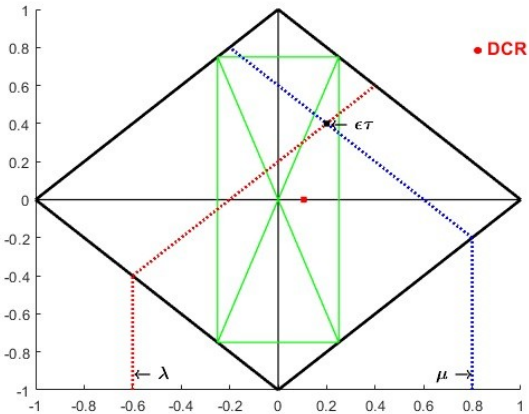

(b)

**Figure 5:** Effects of the $Ft_C$ on the areas of the 12 logical states. Lower t and F areas for higher $Ft_C$ values than 0.5 (a) and lower $\top$ and $\bot$ areas for $Ft_C$ lower than 0.5 (b).

**Table 1:** Fields and terms of Paraconsistent-Lib.

| Field | Symbol | Format | Range | Type | Description |
|---|---|---|---|---|---|
| mu | $\mu$ | float | [0,1] | Input | favorable evidency level |
| lam | $\lambda$ | float | [0,1] | Input | unfavorable evidency |
| FtC | $Ft_C$ | float | [0,1] | Input | Control Factor parameter |
| dc | $D_C$ | float | [-1,1] | Output | certainty degree |
| dct | $D_{CT}$ | float | [-1,1] | Output | contradiction degree |
| d | $d$ | float | [0,√2] | Output | line segment d |
| D | $D$ | float | [0,1] | Output | d limited between [0,1] |
| dcr | $D_{CR}$ | float | [-1,1] | Output | real certainty degree |
| phi | $\varphi$ | float | [0,1] | Output | certainty interval |
| phiE | $\varphi_E$ | float | [0,1] | Output | real certainty interval |
| muE | $\mu_E$ | float | [0,1] | Output | normalized dc or evidence degree |
| muECT | $\mu_{ECT}$ | float | [0,1] | Output | normalized dct or contradition evidence degree |
| muER | $\mu_{ER}$ | float | [0,1] | Output | normalized dcr or real evidence degree |
| decision_output | $S_i$ | float | [0,0.5,1] | Output | 3-state binary decision (V, undefined, F) |
| label | label | str | - | Output | Logical region label |
| regions | Regions | dict | - | Output | Boolean flags per region |

Note: The 12 logical region label are: "t", "f", "⊤", "⊥", "Q⊤→t", "Q⊤→f", "Qt→⊤", "Qf→⊤", "Qt→⊥", "Qf→⊥", "Q⊥→t", "Q⊥→f", "I".

## 4. Applications

In this section we present some examples of how to use the Paraconsistent-Lib and applications.

### 4.1. Testing the Library

The following code is a simple demonstration of how to use the Paraconsistent-Lib. After importing the library, a PAL2v block called "b" is created. Then we set the $Ft_C$, $\mu$ and $\lambda$ values. subsequently, all output variables are displayed, or just one of them ($\mu_{ER}$ in this case).

```
# Testing the Library

# 1) Importing the PAL2v Library
from paraconsistent import ParaconsistentBlock
import numpy as np

# 2) Creating the block
b = ParaconsistentBlock()

# 3) Adjusting Certainty Factor (FtC)
# Values between (0,1)
# It changes the PAL2v 12 states area in the Lattice
# Default value: 0.5
b.config.FtC = 0.5

# 4) Defining inputs mu and lambda
b.input.mu  = 0.70
b.input.lam = 0.60

# 5) Print all outputs
print("=========================")
print("B1:")
b.print_complete()
print("=========================")

# 6) Print also just real certainty evidence (mueR)
print("B1 muER:")
print(b.complete.muER)
```

Fig. 6 present the results. Note that output "Regions" shows 11 logical states as "False" and only one as "True", while "label" output presents only the true state (Q⊤→t in this case).

```
========================
B1:
D: 0.9487
FtC: 0.5000
L: 0.0500
Regions: {'t': False, 'f': False, '⊤': False, '⊥': False, 'Q⊤→t': True, 'Q⊤→f':
False, 'Qt→⊤': False, 'Qf→⊤': False, 'Qt→⊥': False, 'Qf→⊥': False, 'Q⊥→t': False,
'Q⊥→f': False, 'I': False}
V1F: 0.5000
V1V: 0.5000
d: 0.9487
dc: 0.1000
dcr: 0.0513
dct: 0.3000
decision_output: 1.0000
label: Q⊤→t
lam: 0.6000
mu: 0.7000
muE: 0.5500
muECT: 0.6500
muER: 0.5257
phi: 0.7000
phiE: 0.7000
========================
```

**Figure 6:** Results of the Paraconsistent-Lib.

## 4.2. $ParaExtr_{CTX}$ Algorithm for Network Delay Estimation

The most common network test is the "ping" command, which sends Internet Control Message Protocol (ICMP) requests and wait for responses. The size and number of messages can be configured. Normally the number of packets sent, received, and lost are informed, as well as the maximum, minimum, and average delay. This information can be used to evaluate the quality of access to servers, the connection service, especially those sensitive to delays such as streaming, and to diagnose network infrastructure in a local area (LAN) or long distance (WAN).

Here we present a direct use of the $ParaExtr_{CTX}$, to present an alternative and more effective metric instead of the simple arithmetic mean, after extracting contradictions from a set of samples. The full algorithm is available in Appendix A.

The algorithm starts requesting the Internet Protocol (IP) address or HyperText Markup Language (HTML) link, the number, and size of packets. The default values are the google address (www.google.com), 10 packets of 500 Bytes each. Them, the algorithm normalize the delay values, run the $ParaExtr_{CTX}$, denormalizes, and presents the results, as the example in Fig. 7 for the Cloudflare Domain Name Server (DNS), from the script running in Google Colab notebook. By extracting contradictions between values, $ParaExtr_{CTX}$ tends to present more conservative results than arithmetic mean.

```
Ping address [www.google.com]: 1.1.1.1
Number of packets[10]: 12
Packets size (Bytes) [500]: 1000
Pinging 1.1.1.1 com 12 packets 1000 Bytes...

Reply from 1.1.1.1, 1028 bytes in 11.28ms
Reply from 1.1.1.1, 1028 bytes in 11.68ms
Reply from 1.1.1.1, 1028 bytes in 11.06ms
Reply from 1.1.1.1, 1028 bytes in 11.09ms
Reply from 1.1.1.1, 1028 bytes in 10.99ms
Reply from 1.1.1.1, 1028 bytes in 10.94ms
Reply from 1.1.1.1, 1028 bytes in 10.97ms
Reply from 1.1.1.1, 1028 bytes in 11.06ms
Reply from 1.1.1.1, 1028 bytes in 11.15ms
Reply from 1.1.1.1, 1028 bytes in 11.47ms
Reply from 1.1.1.1, 1028 bytes in 10.96ms
Reply from 1.1.1.1, 1028 bytes in 11.11ms

Delays (msec): [11.28, 11.68, 11.06, 11.09, 10.99, 10.94, 10.97, 11.06, 11.15,
11.47, 10.96, 11.11]
Arithmetic Mean of the delay: 11.147 msec
Normalized values (μ) [congestion]: [0.4595 1.     0.1622 0.2027 0.0676 0.
 0.0405 0.1622 0.2838 0.7162
 0.027  0.2297]

=== Final Results ===
ParaExtrCTX μER (congestion) = 0.2857
Estimated ParaExtrCTX (msec)  = 11.151
Arithmetc Average Delay (msec) = 11.147
```

**Figure 7:** Results of the $ParaExtr_{CTX}$ Network Delay Estimation.

## 4.3. PANnet For Best Route Selection

Another way to use the Paraconsistent-Lib is to design and build PANnets. Here, inspired by [Da Silva Filho et al., 2021], we present a simple algorithm using the Paraconsistent-Lib to build a PANnet of four cells, analyzing the network parameters and providing the selection of the right route using the decision_output ($S_1$) at the last cell.

The topology of the PANnet is presented in Fig. 8. In this case, a PANnet analyzes five metrics to present one single indicator ($S_1$) of the best route A or B. The weight of the decision ($\mu_{ER4}$) is also provided. If $\mu_{ER4}$ is higher than $Ft_C$, route A is selected. If $\mu_{ER4}$ is lower than $Ft_C$, route B is the best choice. But if $\mu_{ER4}$ is equal to $Ft_C$, it is undefined, so keep the previous selected route.

The five categories are the normalized values of:
1. Complement of Reception jitter ($\mu_1$).
2. Transmission jitter ($\lambda_1$).
3. Complement of average round trip time (RTT) of the route ($\mu_2$).
4. Processing consumption ($\lambda_2$), and
5. Complement of Packet loss ($\mu_3$).

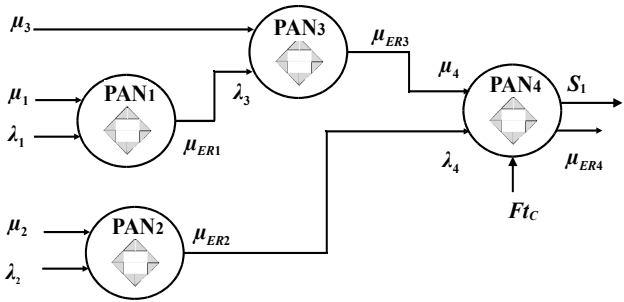

**Figure 8:** Topology of the PANnet for the Paraconsistent Route Selection.

The PANnet for the proposed topology is as simple as the following code.

```
# Create the PANnet
PAN1.input.mu  = mu1
PAN1.input.lam = lam1
PAN2.input.mu = mu2
PAN2.input.lam = lam2
PAN3.input.mu = mu3
PAN3.input.lam = PAN1.complete.muER
PAN4.input.mu = PAN3.complete.muER
PAN4.input.lam = PAN2.complete.muER
```

Appendix B present the entire code. The rest of code is to collect the data, normalize the values and print the results. Fig. 9 presents 3 examples.

```
Reception Jitter (msec): 40
Transmisstion Jitter (msec): 60
round trip time (msec): 50
Processing Consumption of the Router (%): 70
packet loss (%): 20
Result of the PANnet = 0.556
Best Route is = A

Reception Jitter (msec): 10
Transmisstion Jitter (msec): 20
round trip time (msec): 20
Processing Consumption of the Router (%): 95
packet loss (%): 20 Result of the PANnet = 0.500
Best Route is = keep current (undefined analysis)

Reception Jitter (msec): 20
Transmisstion Jitter (msec): 30
round trip time (msec): 40
Processing Consumption of the Router (%): 60
packet loss (%): 40
Result of the PANnet = 0.454
Best Route is = B
```

**Figure 9:** PANnet analysis for the PAL2v Route Selection.

## 5. Conclusion

Paraconsistent-Lib is a novel library that addresses the need of having a simple, practical, open-source, Python API to support the creation of PAL2v reasoning. Users can define just the pair of evidences ($\mu,\lambda$) and one factor of adjustMENT ($Ft_C$) to get three types of outputs. One of the 12 states of the Para-analyser, values of most standard PAN equations including ($D_C$, $D_{CT}$, $D_{CR}$, $\mu_E$, $\mu_{ECT}$, $\mu_{ER}$, $\varphi$) or a decision from $PANC_D$. Paraconsistent-Lib is a valuable addition to the open-source software that supportS PAL2v reasoning, and it is expected to highly facilitate the definition, analysis and interpretation of paraconsistent analysis in a wide variety of AI, decision making, data and knowledge-driven applications. Three examples of applications were also explored in this paper, to encourage researchers who plans to consider PAL2v in their projects to use this novel library, by making the code design easier. In future releases, we plan to extend Paraconsistent-Lib to support additional PAL2v rules and equations to include other types of PAL2v algorithms and PANCs.

**Appendix A**

```python
# Network Test by ParaExtrCTX

# Import Libraries
from pythonping import ping
from paraconsistent import ParaconsistentBlock
import numpy as np
```

```python
# Normalize Inputs
def normalize(values):
    """
    Normalize delays lists to values between [0,1].
    Here, higher the delay, higher the µ value (congestion, traffic delay).
    """
    arr = np.array(values, dtype=float)
    mn, mx = arr.min(), arr.max()
    if mx == mn:
        return np.ones_like(arr), mn, mx
    norm = (arr - mn) / (mx - mn)
    return norm, mn, mx

# Denormalize Results
def denormalize(mu_value, mn, mx):
    """
    denormalize µER for the delay in msec.
    """
    return mn + mu_value * (mx - mn)

# ParaExtrctCTX algorithm
def paraconsistent_reduce(mu_values):
    """
    Apply the Para-extractor algorithm until one last value.
    """
    base = list(mu_values)
    eps = 1e-9

    while len(base) > 1:
        mu = max(base)
        lam_val = min(base)
        lam = 1 - lam_val

        b = ParaconsistentBlock()
        b.input.mu = mu
        b.input.lam = lam

        result = b.complete.muER
        base_new = list(base)  # Create a copy

        # Remove the maximum value from the dataset
        for i, val in enumerate(base_new):
            if abs(val - mu) <= eps:
                base_new.pop(i)
                break

        # Remove the minimum value from the dataset
        for i, val in enumerate(base_new):
            if abs(val - lam_val) <= eps:
                base_new.pop(i)
```

```python
            break

        # Add muER to the dataset
        base_new.append(result)
        base = base_new

    return base[0]

# Execute Ping Sequence
def run_ping_test(host="www.google.com", count=10, size=500):
    """Execute the ping test and storage the results"""
    print(f"Pinging {host} com {count} packets {size} Bytes...\n")
    response = ping(host, count=count, size=size, verbose=True)
    delays = [r.time_elapsed_ms for r in response._responses if r.success]
    if not delays:
        raise RuntimeError("No ping successful!")

    avg_delay = np.mean(delays)
    print("\nDelays (msec):", np.round(delays, 3).tolist())
    print(f"Arithmetic Mean of the delay: {avg_delay:.3f} msec")

    mu_values, mn, mx = normalize(delays)
    print(f"Normalized values (µ) [congestion]:", np.round(mu_values, 4))
    return mu_values, mn, mx, avg_delay

# Request ping parameters
def main():
    host = input("Ping address [www.google.com]: ") or "www.google.com"
    count = int(input("Number of packets[10]: ") or 10)
    size = int(input("Packets size (Bytes) [500]: ") or 500)

    mu_values, mn, mx, avg_delay = run_ping_test(host, count, size)
    muer_final = paraconsistent_reduce(mu_values)
    delay_est = denormalize(muer_final, mn, mx)

# Print results
    print("\n=== Final Results ===")
    print(f"ParaExtrCTX µER (congestion) = {muer_final:.4f}")
    print(f"Estimated ParaExtrCTX (msc)  = {delay_est:.3f}")
      print(f"Arithmetc Average Delay (msec) = {avg_delay:.3f}")

if __name__ == "__main__":
    main()
```

**Appendix B**

```python
# Paraconsistent Analysis Network (PANnet) for Best Route Selection

# Call Paraconsistent Library
from paraconsistent import ParaconsistentBlock
import numpy as np
```

```python
# 1) Create the blocks
PAN1 = ParaconsistentBlock()
PAN2 = ParaconsistentBlock()
PAN3 = ParaconsistentBlock()
PAN4 = ParaconsistentBlock()

# 2) Adjust FtC parameter for all nodes
PAN1.config.FtC = 0.5
PAN2.config.FtC = 0.5
PAN3.config.FtC = 0.5
PAN4.config.FtC = 0.5 # used as threshold for the Route Selection!!!!

# 3) Enter with the data inputs
Rxj = float(input("Reception Jitter (msec): ") or 10)
Txj = float(input("Transmisstion Jitter (msec): ") or 10)
RTT = float(input("round trip time (msec): ") or 10)
PC = float(input("Processing Consumption of the Router (%): ") or 50)
PL = float(input("packet loss (%): ") or 10)

# 3) Normalization and evidence inputs
mu1 = 1 - (Rxj - 0.001)/(100 - 0.1) # Maximum jitter 100ms an Minimum 0.001ms
lam1 = (Txj - 0.001)/(100 - 0.1) # Maximum jitter 100ms an Minimum 0.001ms
mu2 = 1 - (RTT - 0.001)/(200 - 0.001) # Maximum RTT of 200ms and Minimum 0.001ms
lam2 = PC/100 # normalization of %
mu3 = 1 - (PL/100) # normalization of %

# Create the PANnet
PAN1.input.mu  = mu1
PAN1.input.lam = lam1
PAN2.input.mu = mu2
PAN2.input.lam = lam2
PAN3.input.mu = mu3
PAN3.input.lam = PAN1.complete.muER
PAN4.input.mu = PAN3.complete.muER
PAN4.input.lam = PAN2.complete.muER

# Print Results
print(f"Result of the PANnet = {PAN4.complete.muER:.3f}")

if PAN4.complete.decision_output == 1:
 print(f"Best Route is = A")
elif PAN4.complete.decision_output == 0:
 print(f"Best Route is = B")
else:
 print(f"Best Route is = keep current (undefined analysis)")
```